\documentclass[letterpaper, 10 pt, conference]{ieeeconf}  % Comment this line out if you need a4paper
\usepackage{graphicx}
\usepackage{multirow}
\usepackage{xcolor}
\usepackage{amsmath,amssymb}
\usepackage{cite}
\usepackage{caption}

\usepackage{tabularx}
\usepackage{booktabs}
\usepackage[colorlinks,linkcolor=red,anchorcolor=blue,citecolor=green]{hyperref}

\IEEEoverridecommandlockouts                       
\title{\LARGE \bf
CAO-RONet: A Robust 4D Radar Odometry with Exploring More Information from Low-Quality Points
}

\author{Zhiheng Li, Yubo Cui, Ningyuan Huang, Chenglin Pang and Zheng Fang*
\thanks{This work was supported by the National Natural Science Foundation of China under Grants 62073066, the Fundamental Research Funds for Central Universities under Grant N2226001, and 111 Project under Grant B16009. (Corresponding author: Zheng Fang, e-mail: fangzheng@mail.neu.edu.cn)}
\thanks{The authors are all with the Faculty of Robot Science and Engineering, Northeastern University, Shenyang, China; 
}
}

\begin{document}

\maketitle
\thispagestyle{empty}
\pagestyle{empty}

%%%%%%%%%%%%%%%%%%%%%%%%%%%%%%%%%%%%%%%%%%%%%%%%%%%%%%%%%%%%%%%%%%%%%%%%%%%%%%%%
\begin{abstract}
Recently, 4D millimetre-wave radar exhibits more stable perception ability than LiDAR and camera under adverse conditions (e.g. rain and fog).
However, low-quality radar points hinder its application, especially the odometry task that requires a dense and accurate matching.
To fully explore the potential of 4D radar, we introduce a learning-based odometry framework, enabling robust ego-motion estimation from finite and uncertain geometry information. First, for sparse radar points, we propose a local completion to supplement missing structures and provide denser guideline for aligning two frames. Then, a context-aware association with a hierarchical structure flexibly matches points of different scales aided by feature similarity, and improves local matching consistency through correlation balancing. Finally, we present a window-based optimizer that uses historical priors to establish a coupling state estimation and correct errors of inter-frame matching. 
The superiority of our algorithm is confirmed on View-of-Delft dataset, achieving around a 50\% performance improvement over previous approaches and delivering accuracy on par with LiDAR odometry. Our code will be available.
\end{abstract}

%%%%%%%%%%%%%%%%%%%%%%%%%%%%%%%%%%%%%%%%%%%%%%%%%%%%%%%%%%%%%%%%%%%%%%%%%%%%%%%%
\section{INTRODUCTION}
Odometry estimation is an important issue for autonomous driving and mobile robots, aiming to supply precise location information for navigation through correlating sensor data at various times. 
In recent years, some algorithms~\cite{Pwclo, UnsupervisedVO, Translo, SelfVO, SelfLO} have devoted to applying supervised or self-supervised learning to address odometry estimation problem, achieving results that approach or surpass traditional geometry-based methods~\cite{loam,GICP,NDT}.
However, most odometry methods often rely on camera or LiDAR that are easily affected by lighting and weather, making them difficult to handle complex application scenes, such as rain, fog and smoke-filled underground mines.

Due to its high penetration and long-range sensing ability, 4D radar has garnered significant attention and is widely used in perception tasks like 3D detection and tracking~\cite{4DDetection, LXL, Radartracker, Rfast, Rframework, Rcfusion, Smurf}. 
Similarly, several works~\cite{CMFlow,4DRONet,4DRVONet,SelfRONet} also have tried to achieve end-to-end 4D radar odometry to deal with harsh conditions. However, most of them adhere to LiDAR-based paradigm~\cite{Pwclo} without fully considering the distinctive features of radar.
For example, they usually match adjacent raw radar frames based on geometric distance relationships. Subsequently, the inter-frame registration results, which do not account for long-term motion pattern, are directly used as the predicted ego-motion.

Specifically, most algorithms neglect the following issues: (1) \textit{Sparsity}: Each radar frame generally contains only a few hundred points (about 1\% of LiDAR), thus it provides limited geometric information for matching. (2) \textit{Noise}: The position of radar points suffers from noise owing to multipath effects, leading to a ``hard'' distance-based matching struggle to learn reliable data association. (3) \textit{Continuity}: The state estimation viewed as mere inter-frame matching disrupts the continuity of ego-motion and aggravates the impact of error in degraded scenarios, such as being occluded, where finding enough and effective matching pairs is harder for radar than LiDAR.

\begin{figure}[t]
\centering
\includegraphics[width=\linewidth, height=5.0cm]{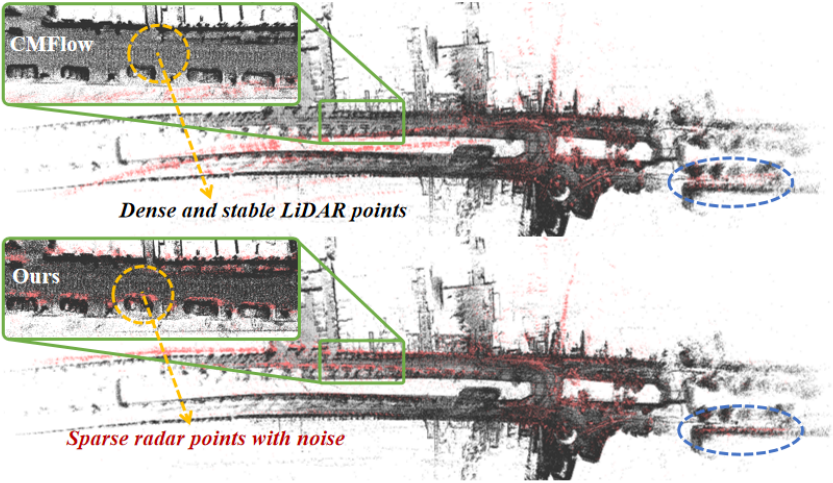}
\caption{Comparison of odometry accuracy. The black points are LiDAR map constructed using ground-truth poses, while red points denote radar map assembled from predicted poses.}
\vspace{-0.2in}
\label{fig:Mapping}
\end{figure} 

To solve the above challenges, we propose a deep-learning odometry network named CAO-RONet, designed to be compatible with the unique properties of 4D radar, which consists of three essential ideas: \textbf{(1) \textit{Local completion}}: For sparse and incomplete radar points, an intuitive strategy is to fill empty spaces. Thus, we create many synthetic points that align with regional structure and supply more geometric information for matching process (e.g., yielding 64 synthetic points from 256 raw points supplies 25\% additional data). By doing so, denser point pairs can be used to reduce odometry errors, especially in turning.
\textbf{(2) \textit{Context-aware association}}: Instead of directly matching points based on distance, we extra consider feature similarity that implicitly compares partial structure and radial velocity. Then, it is softly combined with the distance weight to achieve a resilience registration and alleviate the influence of noise in point positions. Moreover, a sequential modeling is utilized to balance matching information along the multiple directions to suppress outliers and ensure correlation consistency in local area.
\textbf{(3) \textit{Clip-window optimization}}: Following a core principle that odometry is continuous state estimation, we unite the notion of window optimizer with the state space model (SSM) to analyze motion patterns cross state sequence and allow historical priors to restrict the pose estimation. This forms a coupling relationship that rectifies poor matching and smooths trajectory in scenes with insufficient matched points.

In sum, the contributions of our paper are as follows:
\begin{itemize}
    \item We design a 4D radar odometry network, named \underline{CAO}-RONet, to unleash the power of low-quality radar points and implement robust ego-motion estimation over time.
    
    \item We first introduce a local \underline{C}ompletion to provide denser constraint for matching. A context-aware \underline{A}ssociation is adopted to flexibly match points with noise and suppress outliers. We also present a clip-window \underline{O}ptimization to couple multiple states across time to correct pose errors.
    
    \item Extensive experiments on View-of-Delft dataset demonstrate that the proposed method achieves state-of-the-art performance with around a 50\% reduction in root mean square error against previous works, running at 50 FPS.
\end{itemize}

\section{RELATED WORK}
\subsection{LiDAR-based Odometry Methods}
As a classic algorithm, ICP is widely utilized in traditional LiDAR odometry. It strives to align the points of two frames by searching their corresponding relationship and minimizing the distance errors.
Based on error measurement, ICP can be categorized into P2P-ICP~\cite{P2P-ICP} and P2Pl-ICP~\cite{P2Pl-ICP}, aiming to shorten point-to-point and point-to-plane distances. GICP~\cite{GICP} further intends to combine the advantages of both. Compared to ICP with static assumptions, NDT~\cite{NDT} converts point cloud into probability distributions, exhibiting stronger adaptability in dynamic scenes, but determining all points associations is too time-consuming.
Thus, to represent raw points with fewer elements, LOAM~\cite{loam} selects keypoints from sharp edges and planar surfaces based on curvature, and exploits them to align edge lines and planar patches. Then, Lego-LOAM~\cite{lego-loam} uses a segmentation module to discard unreliable points and apply planes derived from stable ground points for matching.

Thanks to powerful data encoding and association abilities of neural networks, end-to-end LiDAR odometry has rapidly developed. LONet~\cite{Lo-net} eases the effect of dynamic objects by a probability mask and constrains network learning through differences in normal vectors between two frames. After that, LodoNet~\cite{Lodonet} uses image-based feature descriptors to extract keypoint pairs from LiDAR images and match them for pose estimation. 
Drawing on the idea of ICP iterative optimization, PWCLO-Net~\cite{Pwclo} proposes a coarse-to-fine strategy to achieve ego-motion refinement utilizing multiple warps. TransLO~\cite{Translo} further applies a window-based Transformer to extract global embeddings for large-scale matching consistency.
While the above works get encouraging results with LiDAR in standard conditions, they still face challenges when point degradation occurs due to heavy smoke or adverse weather.

\subsection{4D Radar-based Odometry Methods}
Due to all-weather operational characteristics, some methods try to use 4D radar to implement odometry. For example, regarding the issue of radar frames losing obvious geometric structure, 4DRadarSLAM~\cite{4dradarslam} integrates GICP with spatial probability distribution of each point and develops APDGICP algorithm for scan-to-scan matching.
Besides, to alleviate the sparsity of points, \cite{4dego-velocity} employs sliding window to construct a dense radar submap with rich structural information, which is aligned with current frame by NDT~\cite{NDT} for scan-to-submap registration.
%%%%%%%%%%%%
As the initial learning-based 4D radar odometry, SelfRO~\cite{SelfRONet} presents a self-supervised method that employs consistency losses based on velocity, geometry and distribution to minimize the gap between two frames.
4DRONet~\cite{4DRONet} decouples radar information encoding with different natures to avoid mutual interference and adopts a velocity-aware cost volume to enable stable matching, even with moving objects.
To diminish reliance on costly labels, CMFlow~\cite{CMFlow} proposes an elaborate cross-modal method that utilizes complementary supervision signals from multi-sensor and other pre-trained models to guide network training.
However, these end-to-end works disregard the sparsity and noise of 4D radar, thus the potential of radar odometry has not been fully explored.

\begin{figure*}[t!]
\centering
\includegraphics[width=\linewidth]{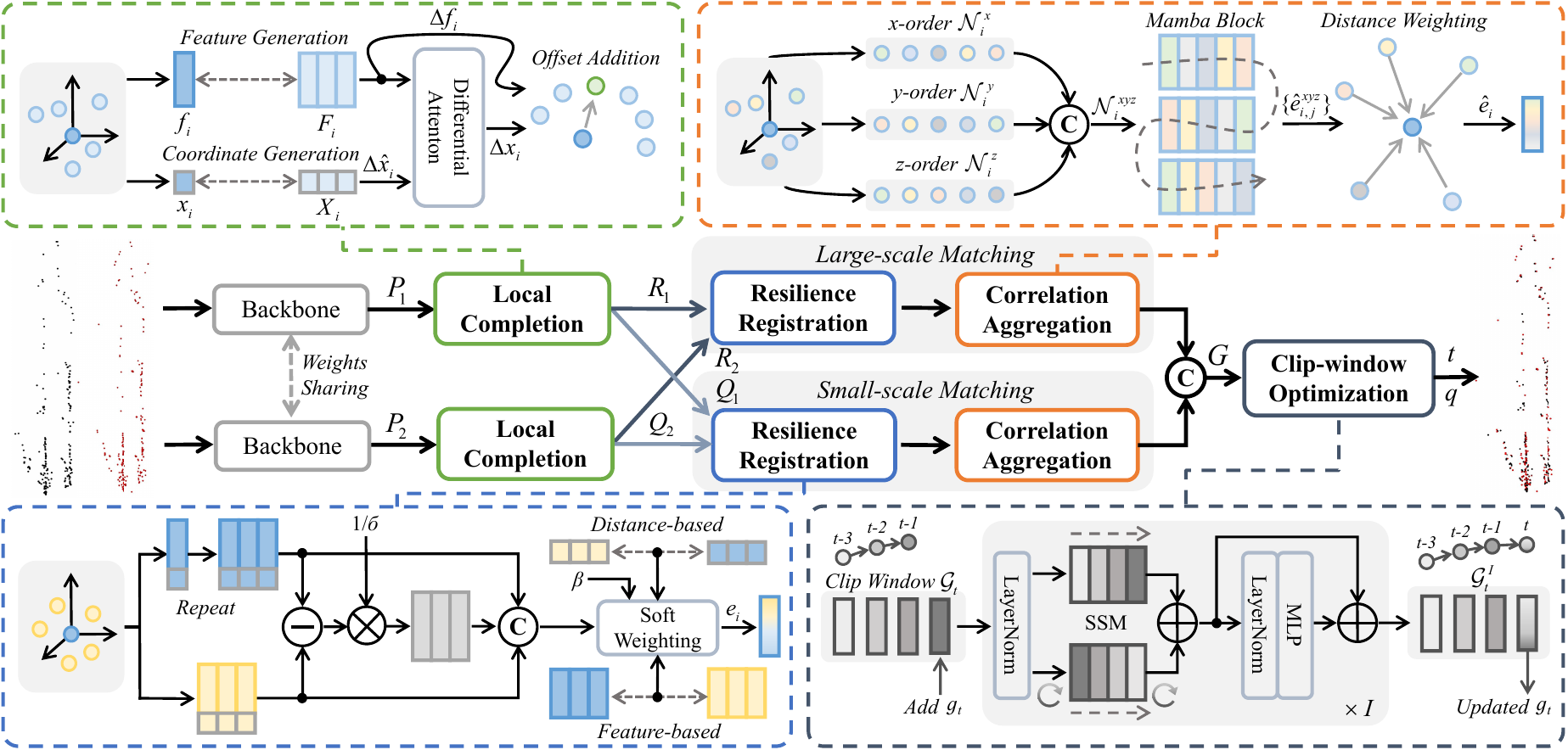}
\caption{The overview of our proposed CAO-RONet. At first, the two frames of radar features derived from backbone are fed into LCM to densify sparse points. Then, CAM implements feature-assisted registration to associate point pairs in different scales, followed by correlation balancing to suppress outliers. Finally, COM with sequential state modeling applies historical prior from clip window to constraint the current ego-motion prediction and smooth trajectory.
}
\vspace{-0.18in}
\label{fig:framework}
\end{figure*} 

\section{METHODOLOGY}
\subsection{Overview} 
Fig.~\ref{fig:framework} illustrates the framework of our CAO-RONet. Two sampled adjacent frames are first input into PointNet++~\cite{Pointnet++} to encode radar information (3D position, radar cross section (RCS), and radial relative velocity (RRV)), resulting in $P_1 = \{p_i = \{x_i, f_i\}\}^N_{i=1}$ and $P_2 = \{p_j = \{x_j, f_j\}\}^N_{j=1}$, where $x \in \mathbb{R}^{3}$ denotes 3D coordinates, $f \in \mathbb{R}^{C}$ is point features. 
%%%
To enhance sparse points, a Local Completion Module (LCM) is proposed to offset $M$ anchor points and generate artificial points, which are combined with $P$ to get densified sets $Q_1 = \{q_i\}^{M+N}_{i=1}$ and $Q_2 = \{q_j\}^{M+N}_{j=1}$.
%%%
Thereafter, the point sets are split into two groups and separately undergo Context-aware Association Module (CAM) to match point pairs by feature-assisted aligning and correlation balancing, yielding a feature $G \in \mathbb{R}^{(M+N+W) \times C}$ with multi-scale matching information.
%%%
Finally, we use a Clip-window Optimization Module (COM) to establish a coupling odometry, which adopts bi-directional SSM to optimize state feature $g_t$ derived from $G$ and estimate a quaternion $q \in \mathbb{R}^4$ and translation vector $t \in \mathbb{R}^3$.

\subsection{Offset-based Local Completion} 
\label{sec:LCM}
Although some algorithms~\cite{Anchorformer,Seedformer,SA-Net,Pointr} are dedicated to point cloud completion, they often focus on repairing missing parts of objects. Thus, directly applying them to sparse radar points from large-scale scenes may cause uncontrolled positions of artificial points and hinder stable local matching.
Rather than completing the scene globally, we employ a local scheme that encodes local information of raw points to predict coordinate and feature offsets, creating new ones with confined positions that conform to regional shape and facilitate denser matching.

Specifically, for point clouds $P \in \{P_1, P_2\}$, we adopt the farthest point sampling (FPS) to determine $M$ anchor points $S=\{s_i=\{x_i, f_i\}\}^M_{i=1}$. Based on each $s_i$, we use ball query to search for $K$ neighbors from $P$ and produce a local region $\mathcal{N}_i$, where the point coordinates and features are $X_{i} \in \mathbb{R}^{K\times 3}$ and $F_{i} \in \mathbb{R}^{K\times C}$. Then, the max-pooling is applied to process $F_i$ and generate a local feature $\overline{f}_i\in \mathbb{R}^{C}$, which is adopted to calculate the difference with $f_i$ and mapped to feature offset $\Delta f_i$ through multi-layer perceptron (MLP):
\begin{equation}
\overline{f}_i = \text{MaxPool}(F_{i}), \ \Delta f_{i} = \text{MLP}(f_{i} - \overline{f}_i)
\end{equation}

Similar to the above procedure, we average the coordinates $X_i$ within local region $\mathcal{N}_i$ to get a geometric center $\overline{x}_i$. Next, the initial positional offset $\Delta \hat{x}_i$ is derived by calculating the difference between $x_i$ and $\overline{x}_i$:
\begin{equation}
\overline{x}_i = \text{AvgPool}(X_i), \ \Delta \hat{x}_i = (x_{i} - \overline{x}_i)
\end{equation}

To further correct initial offset $\Delta \hat{x}_i$ in a learnable manner, we introduce a differential attention mechanism in which the feature $(\Delta f_{i}, f_i)$ is normalized and projected into $\textit{query}$ and $\textit{key}$ embeddings. Then, attention weight is computed by dot-product and exerted on \textit{value} $\Delta \hat{x}_i$ to get updated offset $\Delta x_i$:
\begin{equation}
\Delta x_{i}=\text{Softmax}(\frac{(\text{LN}({\Delta f_{i}}){W_q}) (\text{LN}({f_i}){W_k})^{T}}{\sqrt{{d}_k}}) \Delta \hat{x}_i
\end{equation}
where $W_q, W_k$ are linear projection matrices, and LN is layer normalization. The $\Delta f_{i}$ and $\Delta x_{i}$ are added to anchor point $s_i$ to produce a new point $q_i = \{x_i + \Delta x_i , f_i + \Delta f_i \}$. Finally, these new point are combined with the original $P$ to generate $Q_1 = \{q_i\}^{M+N}_{i=1}$ and $Q_2 = \{q_j\}^{M+N}_{j=1}$ for a denser matching. 

\subsection{Hierarchical Context-aware Association}
\label{sec:RMM}
To avoid excessive reliance on local similarity and neglect of large-scale consistency in inter-frame matching, we adopt a hierarchical strategy that divides matching process into two groups: a small-scale ($Q_1 \rightarrow Q_2$) and large-scale (${R}_1 \rightarrow {R}_2$) registration, where ${R}_1$ and ${R}_2$ are sampled from ${Q}_1$ and ${Q}_2$ using FPS. Subsequently, we apply resilience registration and correlation propagation to each group independently. Due to the same process, we only introduce $(Q_1 \rightarrow Q_2)$ for brevity.

\noindent\textbf{Resilience Registration.}
For the point $q_i = \{x_i, f_i\} \in \mathbb{R}^{3+C}$ in $Q_1$, we find $K$ neighbors $\mathcal{N}_i = \{q_{i,j} = \{x_{i,j}, f_{i,j}\}\}^K_{j=1} \in \mathbb{R}^{K \times (3+C)}$ in $Q_2$. Later, we determine the difference between $q_{i}$ and $q_{i,j}$ and obtain a normalized feature $d_{i,j}$ by scalar $\sigma$, which stands for feature deviations across channels and local groups, thereby reducing the influence of excessive deviation. 
\begin{equation}
d_{i,j} = \frac{q_{i} - q_{i,j}}{\sigma + \epsilon},\sigma = \sqrt{\frac{1}{D} \sum_{i=1}^{M+N} \sum_{j=1}^{K} (q_{i} - q_{i,j}})^2
\label{equ:normalized}
\end{equation}
where $D=(M+N)\times K\times (3+C)$, and $\epsilon$ is a small constant used to ensure numerical stability. Then, the $d_{i,j}$, $q_{i}$ and $q_{i,j}$ are merged and processed through MLP to get a contrastive feature $h_{i,j}$ that measures the relationship between $q_i$ and its neighbor $q_{i,j}$. 
To assemble discrete set $\{h_{i,j}\}^K_{j=1}$ into a local correlation vector $e_i \in \mathbb{R}^{C}$, we produce two types of weights $(w^d_{i,j} \in \mathbb{R}^{C}, w^f_{i,j}\in \mathbb{R}^{C})$ and then apply soft weighted sum to $h_{i,j}$, using a learnable parameter $\beta$ to adjust the confidence in both the spatial distance and feature similarity, as follows:
\begin{equation}
h_{i,j} = \text{MLP}(d_{i,j} \oplus q_{i} \oplus q_{i,j})
\end{equation}
\begin{equation}
w^d_{i,j} = \text{MLP}(x_{i} - x_{i,j}), w^f_{i,j} = \text{MLP}\langle f_{i},  f_{i,j}\rangle
\label{equ:weight}
\end{equation}
\begin{equation}
e_{i} = \beta \sum_{j=1}^{K} h_{i,j} \odot w^f_{i,j} + (1-\beta ) \sum_{j=1}^{K} h_{i,j} \odot w^d_{i,j}
\label{equ:weighted}
\end{equation}
where $\oplus$ and $\odot$ mean channel concatenation and dot product. $\langle \cdot \rangle$ denotes similarity calculation. As a result, after combining vector $e_i$ of each point, we obtain the correlation embedding $E = \{e_i\}^{M+N}_{i=1}$ for $Q_1$, which accounts for feature difference to enable more resilient matching compared to relying solely on rigid distance affected by random noise of point position.

\noindent\textbf{Correlation Aggregation.}
It is worth noting that some points may have insufficient matching due to occlusion or isolation, giving rise to outliers in correlation embedding $E$. A simple way to suppress them is to aggregate the correlations within a local region $\mathcal{N}_i = \{q_{i,j} = \{x_{i,j}, e_{i,j}\}\}^K_{j=1} $ surrounding each point $q_i = \{x_{i}, e_{i}\}$ of $Q_1$ and then update $e_i$ to improve local consistency.
However, this approach may still be affected by outliers in $\mathcal{N}_i$.
To tackle this matter, we advocate sorting the points within $\mathcal{N}_i$ through coordinates $\{x_{i,j}\}$ and applying an RNN-like sequential modeling to adjust each embedding $e_{i,j}$ by neighbors, which is based on the fact that adjacent points exhibit similar matching situations in most cases.

Specifically, we first arrange points $\mathcal{N}_i$ along the $x$, $y$ and $z$ axes through their coordinates and generate three sequences $(\mathcal{N}_i^x, \mathcal{N}_i^y, \mathcal{N}_i^z)$, which are concatenated to obtain $\mathcal{N}_i^{xyz}$ with coordinates $\{x_{i,j}^{xyz}\}^{3K}_{j=1}$ and embeddings $\{e_{i,j}^{xyz}\}^{3K}_{j=1}$.
Then, a Mamba block~\cite{mamba} with a global receptive field and parallel processing is used to encode $\{e_{i,j}^{xyz}\}$ sequentially and produce balanced embeddings $\{\hat{e}_{i,j}^{xyz}\}$.
Later, to aggregate $\{\hat{e}_{i,j}^{xyz}\}$ and refine correlation $e_i$, we calculate Euclidean distance weights $w_{i,j}$ between $q_i$ and its neighbors $\mathcal{N}_i^{xyz}$, which are assigned to $\{\hat{e}_{i,j}^{xyz}\}$ to get a updated correlation embedding $\hat{e}_i$ in Eq.~\ref{equ:Euclidean}.
\vspace{-0.1in}
\begin{equation}
w_{i,j} = \frac{1}{||x_{i} - x_{i,j}^{xyz}||_2}, \hat{e}_i = \text{MLP}(\sum_{j=1}^{3K}(w_{i,j} \odot \hat{e}_{i,j}^{xyz}) \oplus e_i)
\label{equ:Euclidean}
\vspace{-0.07in}
\end{equation}

Finally, we combine the correlation embeddings from both ($Q_1 \rightarrow Q_2$) and ($R_1 \rightarrow R_2$) into feature $G \in \mathbb{R}^{(M+N+W) \times C}$ that contains multi-scale alignment. $W$ is the number of $R_1$.

\subsection{Bi-directional Clip-window Optimization} 
\label{sec:WOM}
Different from previous method~\cite{4DRONet} that pools correlation embedding $G$ to obtain state quantity $g_t \in \mathbb{R}^{1\times C}$ and directly predict ego-motion, we argue that it is necessary to introduce historical states as constraints to handle transient degradation issue. Thus, we construct a clip window of maximum length $L$ to store states. Its update mechanism is denoted as follows:
\begin{equation}
\mathcal{G}_t = \begin{cases} 
\{ g_t \}, & \text{if } t \bmod L = 0 \\
\{ g_{t-(t \bmod L)},..., g_t \}, & \text{otherwise}
\end{cases}
\vspace{-0.02in}
\end{equation}
Specifically, after obtaining the raw current state $g_t$, it is first added to $\mathcal{G}_t$. If clip window is full, the past states are cleared, and the window is refilled to prevent the prolonged influence of low-quality states. To leverage historical priors to optimize $g_t$, we define a discretized state space model (SSM)~\cite{S4} as in Eq.~\ref{equ:SSM}, where the $\overline{\mathbf{A}} \in \mathbb{R}^{C\times C}$, $\overline{\mathbf{B}} \in \mathbb{R}^{C\times 1}$ and $\mathbf{C} \in \mathbb{R}^{1\times C}$ denote learnable parameters. Note that the hidden state $h_{t-1}$ that implicitly represents motion pattern is derived from past states of $\mathcal{G}_t$ by the same way, and this process is performed in parallel with a convolution kernel~\cite{SSM} rather than recursion.
\begin{equation}
h_t = \overline{\mathbf{A}}h_{t-1}+\overline{\mathbf{B}}g_t, \ \hat{g}_t = \mathbf{C}h_t
\label{equ:SSM}
\end{equation}
Based on SSM, we further introduce a bidirectional modeling block as shown in Eq.~\ref{equ:Bi-SSM} and Eq.~\ref{equ:FFN}, where ordered sequence $\mathcal{G}_t$ is fed into the SSM in both forward and reverse directions to enable network to learn a broader range of motion patterns.
\begin{equation}
\hat{\mathcal{G}}^i_t = \text{SSM}(\text{LN}(\mathcal{G}_t^{i-1})) + \mathcal{F}( \text{SSM}(\mathcal{F}(\text{LN}(\mathcal{G}_t^{i-1}))))
\label{equ:Bi-SSM}
\end{equation}
\begin{equation}
\mathcal{G}^i_t = \text{MLP}(\text{LN}(\hat{\mathcal{G}}^i_t)) + \hat{\mathcal{G}}^i_t
\label{equ:FFN}
\end{equation}
where $i$ and $\mathcal{F}$ denote $i$-th block ($i \in \{1,...,I\}$) and reverse sorting. To the end, we separate updated $g_t$ from final output $\mathcal{G}_t^I$ and use two MLPs to estimate the quaternion $q \in \mathbb{R}^4$ and translation vector $t \in \mathbb{R}^3$.

\begin{figure}[t!]
\centering
\vspace{0.1in}
\includegraphics[width=\linewidth, height=3.85cm]{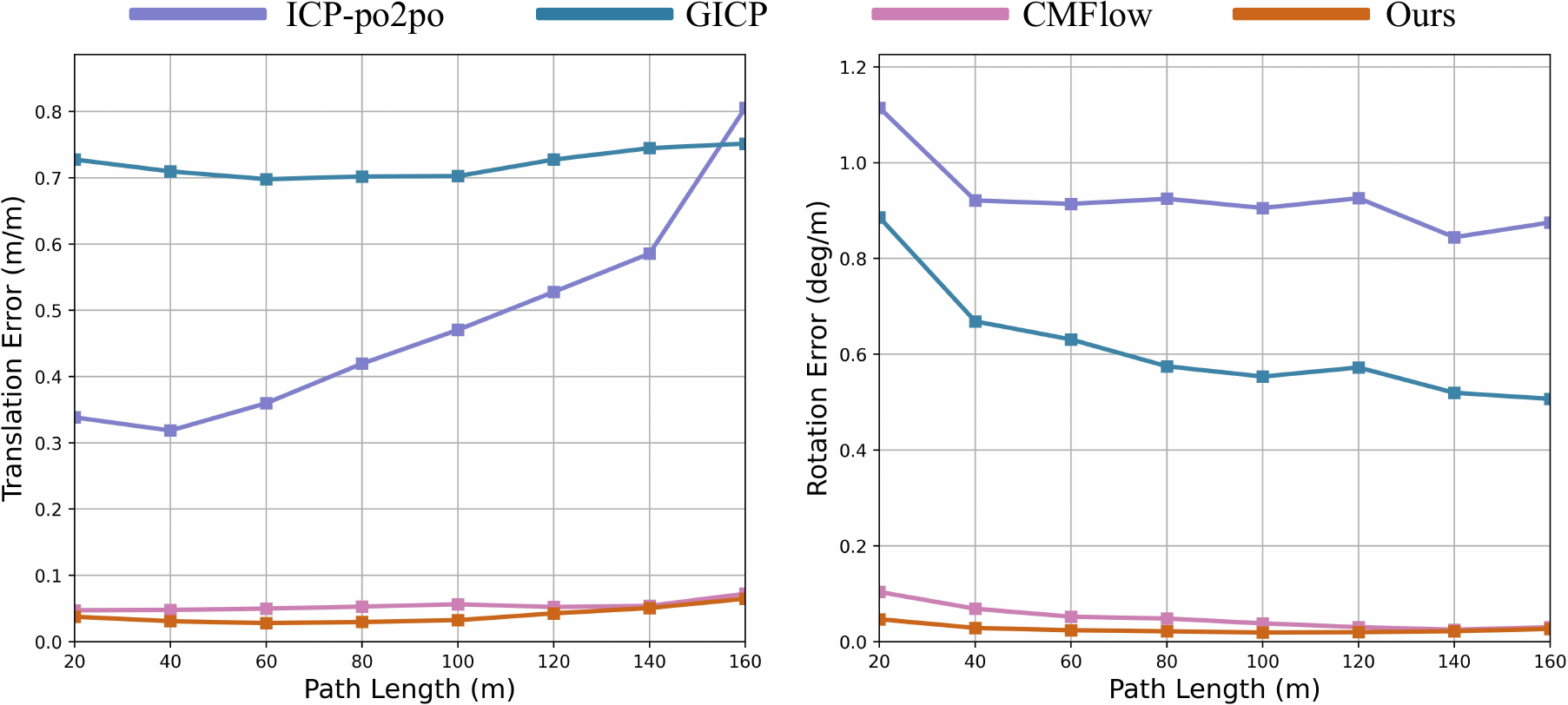}
\caption{Average translational and rotational errors on the test sequences of VoD in the length of 20, 40, ..., 160$m$.}
\label{fig:Error}
\vspace{-0.2in}
\end{figure} 

\subsection{Implementation Details}
\noindent\textbf{Loss Functions.}
The loss functions for rotation and translation are defined in Eq.~\ref{equ:loss}, where $q^{gt}$ and $t^{gt}$ are the ground-truth quaternion and translation vector, respectively.
\begin{equation}
\mathcal{L}_q = || q - q^{gt} ||_2, \mathcal{L}_t = || t - t^{gt} ||_2 
\label{equ:loss}
\end{equation}

Referring to the previous work~\cite{4DRONet}, we use two learnable parameters, $w_q$ and $w_t$, which adjust losses $\mathcal{L}_q$ and $\mathcal{L}_t$ during training to account for differences in scale and units between $q$ and $t$. Consequently, the overall loss can be formulated as:
\begin{equation}
\mathcal{L} = \mathcal{L}_q \text{exp}(-w_q) + w_q + \mathcal{L}_t \text{exp}(-w_t) + w_t
\end{equation}

\noindent\textbf{Data Augmentation.} In data processing, we clear out radar points outside the field of view of image and constrain them to the height range of [-3$m$, 3$m$] to retain reliable points. To increase data diversity, we flip training sequences to produce new trajectories with reverse ego-motion, and apply random offsets to points and ground-truth pose matrix during training

\noindent\textbf{Training \& Inference.}
The model is trained for 60 epochs on a NVIDIA RTX 4090 GPU using an Adam optimizer with a starting learning rate of 1 $\times$ 10$^{-3}$ and a decay rate of 0.9 per epoch. Then, trainable parameters $w_q$ and $w_t$ are initialized to -2.5 and 0.0, and the clip window length is set to 5. The default numbers of sampled raw points $N$, completion points $M$ and large-scale points $W$ are 256, 64 and 64.

\begin{table*}
\renewcommand\tabcolsep{7.0pt}
\caption{4D radar odometry experiment results on View-of-Delft (VoD) dataset. Like previous works, we keep two decimal places for odometry metric in this table, but three decimal places for ablation studies to better illustrate performance changes.}
\vspace{-0.1in}
\begin{center}
\scalebox{0.9}{
\begin{tabular}{cc|cc|cc|cc|cc|cc|cc|c}
\toprule[.05cm]
\multicolumn{2}{c|}{\multirow{2}{*}{Method}} & \multicolumn{2}{c|}{00} & \multicolumn{2}{c|}{03} & \multicolumn{2}{c|}{04} & \multicolumn{2}{c|}{07} & \multicolumn{2}{c|}{23} & \multicolumn{2}{c|}{Mean} & \multicolumn{1}{c}{\multirow{2}{*}{Time ($ms$)}} \\ \cline{3-14} 
\multicolumn{2}{c|}{} & $t_{rel}$ & $r_{rel}$ & $t_{rel}$ & $r_{rel}$ & $t_{rel}$ & $r_{rel}$ & $t_{rel}$ & $r_{rel}$ & $t_{rel}$ & $r_{rel}$ & $t_{rel}$ & $r_{rel}$  \\ \hline\hline
\multicolumn{1}{c|}{\multirow{4}{*}{Classical-based}} & ICP-po2po & 0.57 & 1.23 & 0.38 & 0.98 & 0.21 & 1.15 & 0.30 & 1.75 & 0.18 & 0.49 & 0.33 & 1.12 & 3.80         \\
\multicolumn{1}{c|}{} & ICP-po2pl & 0.69 & 1.67 & 0.41 & 2.16 & 0.39 & 1.86 & 0.74 & 2.77 & 1.38 & 1.07 & 0.72 & 1.91 & 1.11   \\
\multicolumn{1}{c|}{} & GICP & 0.41 & 0.42 & 0.46 & 0.65 & 0.31 & 0.38 & 0.37 & 0.29 & 0.79 & 0.17 & 0.47 & 0.38 & 1.29    \\
\multicolumn{1}{c|}{} & NDT & 0.52 & 0.63 & 0.56 & 1.52 & 0.47 & 0.91 & 0.69 & 0.51 & 0.52 & 0.37 & 0.55 & 0.79 & 1.02  \\ \hline
\multicolumn{1}{c|}{\multirow{2}{*}{LiDAR-based}} & A-LOAM w/o mapping & - & - & - & - & 0.14 & 0.35 & 0.13 & 0.74 & 0.25 & 1.39 & - & -  & 4.70      \\
\multicolumn{1}{c|}{} & LO-Net & 0.81 & 0.81 & 1.12 & 1.89 & 0.23 & 0.46 & 0.19 & 0.21 & 0.53 & 1.07 & 0.58 & 0.89 & 11.6 \\ \hline
\multicolumn{1}{c|}{\multirow{4}{*}{4D Radar-based}} & RaFlow & 0.61 & 0.84 & 0.87 & 1.98  & 0.07 & 0.45 & 0.07 & 0.04 & 0.42 & 1.16 & 0.41 & 0.90 & 36.3         \\
\multicolumn{1}{c|}{} & 4DRO-Net & 0.08 & \textbf{0.03} & 0.06 & 0.05 & 0.08 & 0.07 & 0.05 & 0.03 & 0.10 & 0.15 & 0.07 & 0.07 & 10.8 \\
\multicolumn{1}{c|}{} & CMFlow$\dag$ & \textbf{0.04} & 0.05 & 0.07 & 0.09 & 0.06 & 0.09 & 0.03 & 0.04 & 0.09 & 0.14 & 0.06 & 0.08 & 30.4 \\
\multicolumn{1}{c|}{} & \textbf{Ours} & 0.05 & \textbf{0.03} & \textbf{0.02} & \textbf{0.03} & \textbf{0.03} & \textbf{0.05} & \textbf{0.02} & \textbf{0.02} & \textbf{0.04} & \textbf{0.06} & \textbf{0.03} & \textbf{0.04} & 20.2\\
\toprule[.05cm]
\end{tabular}
}
\vspace{-0.18in}
\end{center} 
\label{tab:Quantitative}
\end{table*}

\begin{figure*}[t!]
\centering
\includegraphics[width=\linewidth, height=4.1cm]{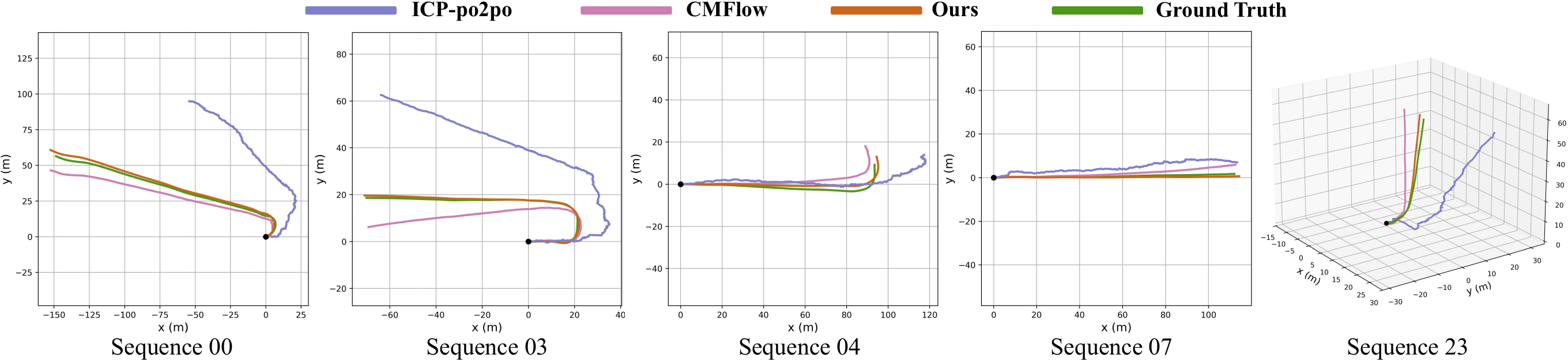}
\caption{The trajectory visualization of our CAO-RONet with other methods on sequences 00, 03, 04, 07 and 23, respectively.}
\vspace{-0.15in}
\label{fig:Trajectory}
\end{figure*} 

%%%%%%%%%%%%%%%%%%%%%%%%%%%%%%%%%%%%%%%%
%%%%%%%%%%%%%%%%%%%%%%%%%%%%%%%%%%%%%%%%
\section{EXPERIMENTS}
\noindent\textbf{Datasets.} We conduct in-depth experiments on View-of-Delft dataset (VoD)~\cite{VoD}, which contains 8,682 point cloud frames (captured by a Velodyne HDL-64 LiDAR and ZF FR-Gen21 4D radar) along with the corresponding extrinsic parameters and odometry information.
Based on the continuity of frame, the VoD can be divided into 24 sequences, with 6,964 frames used as the training set and 1,718 frames from five sequences (00, 03, 04, 07 and 23) used as the test set.

\noindent\textbf{Evaluation Metric.} The relative pose error (RPE) is usually used to measure the difference between the ground truth and predicted poses over the specific intervals or distances. 
Based on it, we calculated the root mean square error (RMSE) for rotation ($^{\circ}/m$) and translation ($m/m$) across test sequences with lengths ranging from 20$m$ to 160$m$ in 20$m$ increments.

\subsection{Quantitative Results.}  
Consistent with~\cite{4DRONet}, our method is compared with classic geometry-based algorithms, such as ICP-po2po, ICP-po2pl, GICP, NDT and LOAM in Tab.~\ref{tab:Quantitative}. Besides, we also compare learning-based methods initially designed for LiDAR points (LO-Net) and 4D radar (RaFlow, 4DRO-Net, and CMFlow).
Specifically, we retrained CMFlow$\dag$ instead of directly using the pre-trained model like~\cite{4DRONet} to avoid unfairness caused by differences in the training sequence split.
The results confirm that classic approaches effective for LiDAR, perform poorly or cannot complete all sequences due to the extremely sparse points of 4D radar. 
Compared to end-to-end methods, owing to improvements made to radar natures, we obtain the lowest mean $t_{rel}$ and $r_{rel}$, and the errors of all sequences are more balanced.
Moreover, Fig.~\ref{fig:Error} shows the average segment errors on test sequences, proving the advantages of our method over previous works.
Finally, we validate that our low-cost radar-only method achieves competitive results compared to other methods that combine camera or use LiDAR in Tab.~\ref{tab:multi-sensor}.

\begin{table}[t!]
\renewcommand\tabcolsep{5.5pt}
\caption{Odometry experiments using different sensors on VoD dataset. Sequence division follows 4DRVO-Net~\cite{4DRVONet}.}
\vspace{-0.1in}
\begin{center}
\scalebox{0.9}{
\begin{tabular}{c|c|c|c|c}
\toprule[.05cm]
Method & \begin{tabular}[c]{@{}c@{}}A-LOAM \\ w/o mapping\end{tabular} & 4DRVO-Net & CMFlow & \textbf{Ours}  \\ \hline\hline
Sensor & LiDAR & Radar + Camera & Radar Only & Radar Only \\
Mean $t_{rel}$ & \textbf{0.06} & 0.08 & 0.11 & \underline{0.07}  \\
Mean $r_{rel}$ & 0.10 & \underline{0.07} & 0.31 & \textbf{0.05} \\
\toprule[.05cm]
\end{tabular}
}
\end{center}
\label{tab:multi-sensor}
\vspace{-0.3in}
\end{table}

\subsection{Qualitative Results}
\begin{figure*}[t!]
\centering
\includegraphics[width=\linewidth, height=5.0cm]{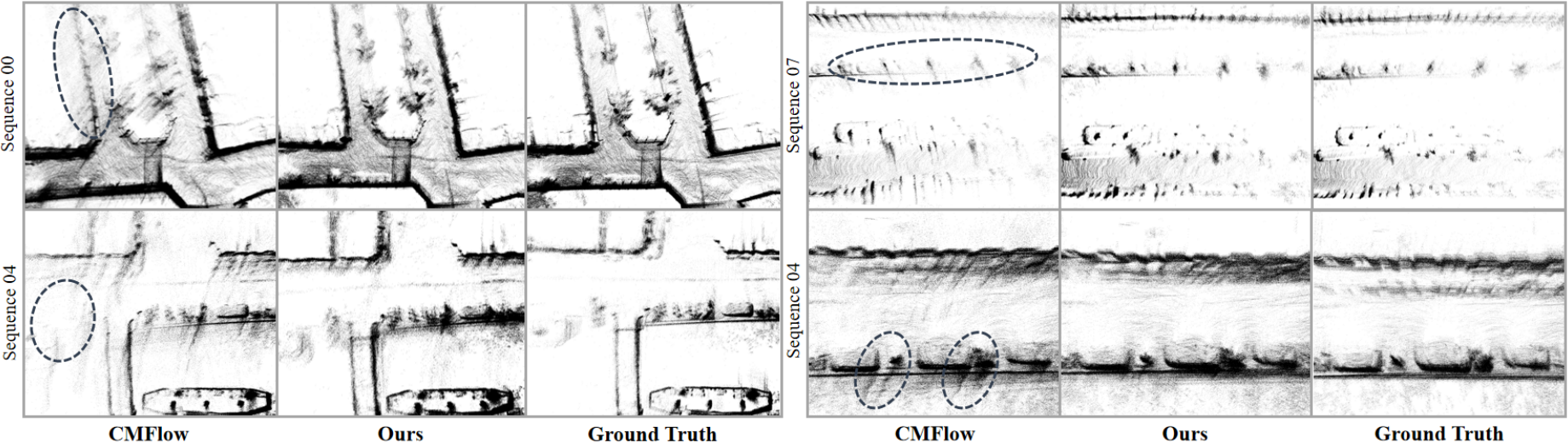}
\caption{Comparison of local maps that are constructed from LiDAR points based on predicted radar odometry.}
\label{fig:Local Mapping}
\vspace{-0.2in}
\end{figure*}

To provide a more intuitive comparison, Fig.~\ref{fig:Trajectory} visualizes the trajectories of several methods across different sequences.
Despite ICP-po2po getting the best $t_{rel}$ among classic methods, it suffers from obvious deviations due to the challenges in building stable matching caused by noise and the sparsity of radar points.
While CMFlow, as a state-of-the-art method, displays considerable improvement, it still produces notable error during turn in sequences 00 and 03.
By contrast, owing to a denser matching and coupling optimization, our method exhibits smaller error during linear and rotational movement, particularly in sequences 03 and 07.
%%%%%%%%%%%
Since the quality of map constructed by odometry can manifest ego-motion accuracy, we apply the predicted poses to LiDAR points to build dense maps that allow for easier comparison. As shown in Fig.~\ref{fig:Local Mapping}, CMFlow produces many ghost effects, whereas our maps are clearer, validating our odometry is more stable and accurate.

\begin{figure}[t]
\centering
\includegraphics[width=8.5cm, height=3.3cm]{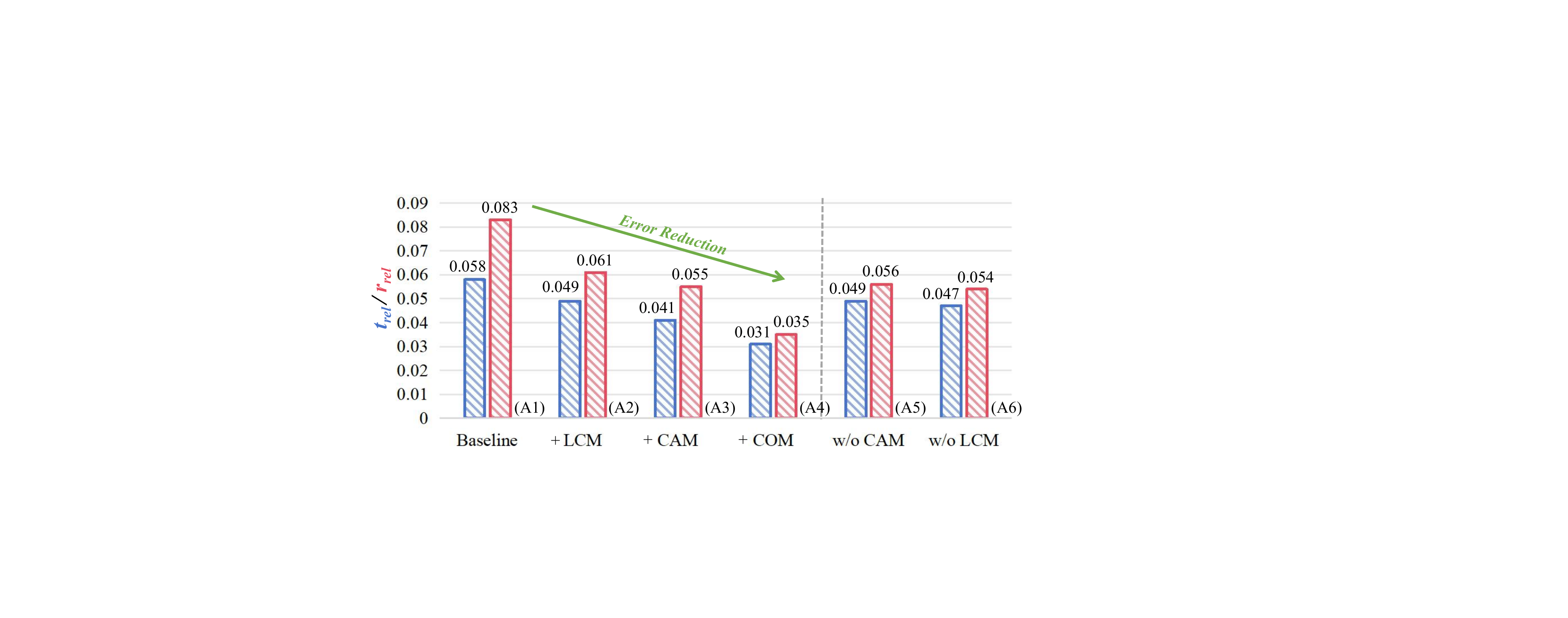}
\caption{The effect of different modules on VoD dataset.}
\vspace{-0.25in}
\label{fig:Modules}
\end{figure} 

\subsection{Ablation Studies}
To thoroughly analyse the impact of each module and its design strategy on our method's performance (mean $t_{rel}$ and $r_{rel}$), we conduct a series of ablation studies on VoD dataset.

\noindent\textbf{Model Components.} 
In Fig.~\ref{fig:Modules}, we can observe that adopting LCM for local completion (A2) results in a moderate reduction in $t_{rel}$ and $r_{rel}$. We attribute this to enhanced geometric information between two frames, which helps network form more effective matching pairs to constrain motion estimation, particularly during turns.
%%%%%%%%
Then, the elastic matching in CAM reduces the impact of noise, while the correlation aggregation balances the matching information of each point to mitigate outlier effects, thereby improving odometry accuracy in A3.
%%%%%%%%
When COM is employed for coupling state optimization, the best result is obtained in A4. Thus, we think that considering historical states is a key factor for the robustness of odometry, especially in degraded situations, where historical priors can help suppress severe errors.
%%%%%%%%
Finally, in A5 and A6, removing CAM or LCM will lead to performance degradation, proving that each module is complementary and indispensable.

\begin{table}[t!]
\renewcommand\tabcolsep{7.0pt}
\caption{Ablation studies on Context-aware Association.}
\label{tab:Relation-aware matching}
\vspace{-0.15in}
\centering
\begin{center}
\scalebox{0.9}{
\begin{tabular}{c|cccc|cc}
\toprule[.05cm]
\multirow{2}{*}{} & \multicolumn{1}{c|}{\multirow{2}{*}{Hierarchy}} & \multicolumn{2}{c|}{Matching}           & \multirow{2}{*}{Aggregation} & \multirow{2}{*}{$t_{rel}$} & \multirow{2}{*}{$r_{rel}$} \\ \cline{3-4}
\rule{0pt}{8pt}
& \multicolumn{1}{l|}{} & Distance & \multicolumn{1}{l|}{Feature} & & & \\ \hline\hline
B1 & & \checkmark &  &  & 0.049 & 0.056 \\
B2 & \checkmark & \checkmark & & & 0.038 & 0.055 \\ 
B3 & & \checkmark & \checkmark & & 0.039 & 0.046 \\ 
B4 & \checkmark & \checkmark & \checkmark & & 0.035 & 0.041 \\
B5 & \checkmark & \checkmark & \checkmark & \checkmark & \textbf{0.031} & \textbf{0.035} \\
\toprule[.05cm]
\end{tabular}
}
\end{center} 
\vspace{-0.2in}
\end{table}

\begin{table}[t!]
\renewcommand\tabcolsep{7.8pt}
\caption{Ablation studies on Clip-window Optimization. Uni- and bi- mean uni-directional and bi-directional SSM.}
\label{tab:optimization}
\vspace{-0.05in}
\centering
\scalebox{0.9}{
\begin{tabular}{c|ccccc|cc}
\toprule[.05cm]
\multirow{2}{*}{} & \multicolumn{1}{c|}{\multirow{2}{*}{Cross-Attn}} & \multicolumn{2}{c|}{SSM} & \multicolumn{1}{c|}{\multirow{2}{*}{Slide}} & \multirow{2}{*}{Clip} & \multirow{2}{*}{$t_{rel}$} & \multirow{2}{*}{$r_{rel}$} \\ \cline{3-4}
\rule{0pt}{8pt}
& \multicolumn{1}{c|}{} & Uni- & \multicolumn{1}{c|}{Bi-} & \multicolumn{1}{c|}{} & & & \\ \hline\hline
C1 & \checkmark & & & & \checkmark & 0.038 & 0.044 \\
C2 & & \checkmark & & & \checkmark & 0.036 & 0.038 \\
C3 & & & \checkmark & \checkmark & & 0.032 & 0.037 \\
C4 & & & \checkmark & & \checkmark & \textbf{0.031} & \textbf{0.035} \\
\toprule[.05cm]
\end{tabular}
}
\vspace{-0.25in}
\end{table}

\begin{figure}[t]
\centering
\includegraphics[width=\linewidth, height=4.3cm]{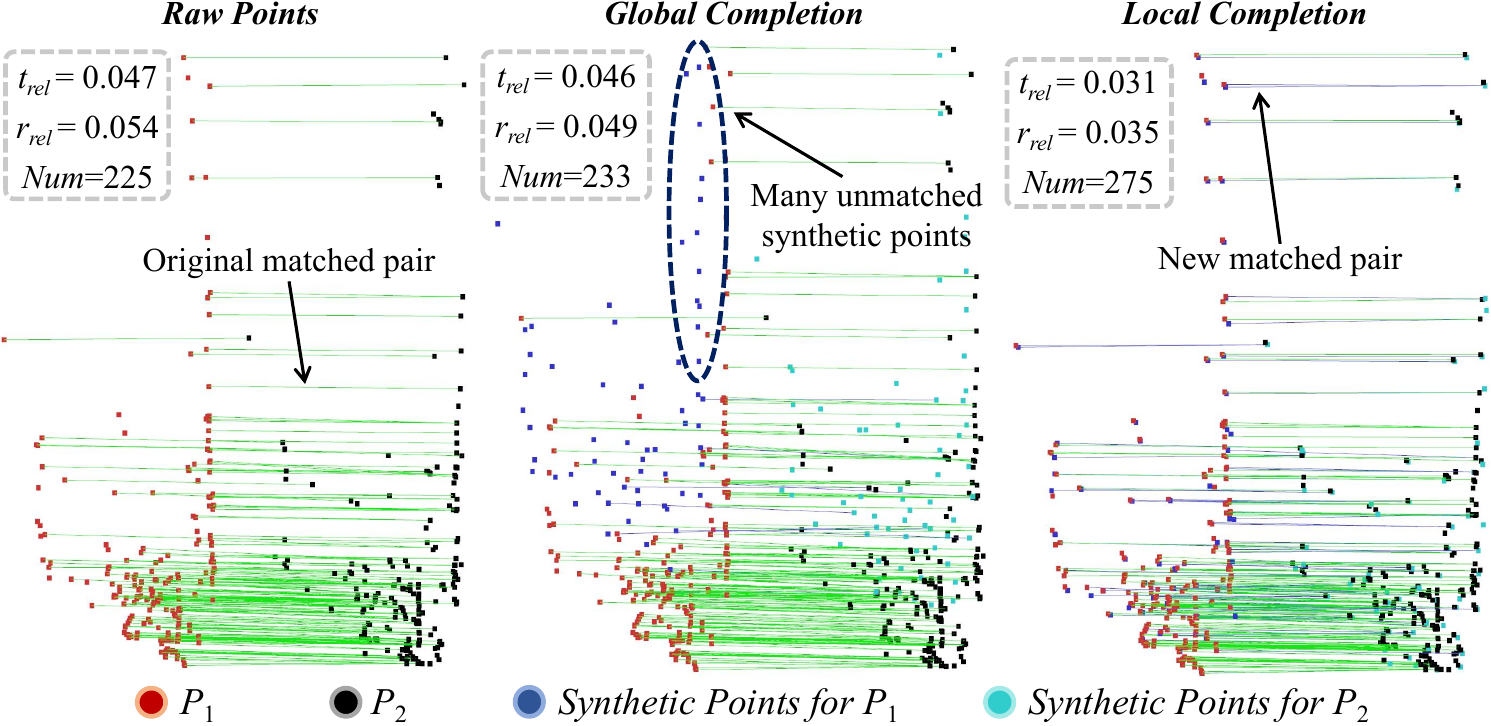}
\caption{Comparison of matching results within a fixed range (1.5$m$) with different completion methods. \textit{Num} is the number of matched pairs. $t_{rel}$ and $r_{rel}$ are mean error of test set.}
\vspace{-0.25in}
\label{fig:Registeration}
\end{figure}

\noindent\textbf{Global Completion vs. Local Completion.}
Previous completion methods~\cite{Anchorformer,Seedformer} usually complete the entire object or scene globally. Although this approach can introduce some additional geometric information, it will result in an unstable distribution of generated points, making it difficult to search correspondence between two frames (in the middle of Fig.~\ref{fig:Registeration}).
In contrast, our local completion restricts point generation to specific areas while adhering to local structural characteristic. Thus, it ensures more consistent created points across frames, enabling denser matching (in the right of Fig.~\ref{fig:Registeration}).

\noindent\textbf{Context-aware Association.} 
We conduct more detailed ablation studies of CAM in Tab.~\ref{tab:Relation-aware matching}. As displayed in B1 and B3, integrating feature-based matching, which takes into account comprehensive information, outperforms the purely distance-based matching due to the unsteady positions of radar points. Then, B2 shows that hierarchical matching encourages model to perceive correspondences at larger scales and contributes to the reduction of $t_{rel}$. Furthermore, B4 and B5 reflect that sequential modeling can ensure adjacent points within a local area share similar correlation information, thereby preventing outliers from misleading ego-motion estimation.

\noindent\textbf{Clip-window Optimization.} 
In order to capture continuous ego-motion, we introduce a state optimizer to save historical state quantities in clip window and use them to update current state. Ablation studies on the design strategy of optimizer are presented in Tab.~\ref{tab:optimization}. In C1, we first try to use cross-attention to establish relationship among multiple states and renew the current one. However, this feature-similarity-based method is challenging to capture the causal and temporal dependencies between states, thus limiting accuracy improvement.
%%%%%%%%
Inspired by global sequence modeling of the state space model (SSM), we adopt it to optimize the current state based on prior states and surpass C1. Later, a bidirectional SSM is designed and achieves a better result in C4 since it models richer motion patterns by flipping state order. 
%%%%%%%%
Finally, owing to clearing the accumulated states within fixed intervals, clip window avoids the sustained influence of inferior states and has advantages over the slide window in our task, as verfied in C3 and C4.

%%%%%%%%%%%%%%%%%%%%%%%%%%%%%%%%%%%%%%%%
%%%%%%%%%%%%%%%%%%%%%%%%%%%%%%%%%%%%%%%%
\section{CONCLUSIONS}
In this article, we analyze the unique characteristics of 4D radar and present an odometry network customized for low-quality radar points. 
It can not only supply denser constraints for matching by local completion but also utilize the feature-assisted registration and correlation balancing to alleviate the impact of noise and outlier. Finally, an ego-motion that aligns with motion trends is estimated by window-based optimizer. 
The experiments show that our method achieves state-of-the-art results against past traditional and learning-based works.

\bibliographystyle{ieeetr}
\bibliography{ref}

\end{document}